%% file: main-eg.tex
\documentclass[conference]{IEEEtran}
\IEEEoverridecommandlockouts

\usepackage{cite}
\usepackage{amsmath,amssymb,amsfonts}
\usepackage{algorithmic}
\usepackage{graphicx}
\usepackage{textcomp}
\usepackage{xcolor}
\usepackage{booktabs}
\usepackage{url}
\makeatletter
\g@addto@macro{\UrlBreaks}{\do\/\do-\do.\do=\do\_\do?\do&\do\%\do\#}
\makeatother
\usepackage{bbm}
\def\BibTeX{{\rm B\kern-.05em{\sc i\kern-.025em b}\kern-.08em
    T\kern-.1667em\lower.7ex\hbox{E}\kern-.125emX}}
\begin{document}

\title{Multi-Agent Large Language Model Based Emotional Detoxification Through Personalized Intensity Control for Consumer Protection}

\author{\IEEEauthorblockN{Keito Inoshita}
\IEEEauthorblockA{\textit{Faculty of Business and Commerce} \\
\textit{Kansai University}\\
Suita, Japan \\
Email: inosita.2865@gmail.com}
}

\maketitle

\begin{abstract}
In the attention economy, sensational content exposes consumers to excessive emotional stimulation, hindering calm decision-making. This study proposes Multi-Agent LLM-based Emotional deToxification (MALLET), a multi-agent information sanitization system consisting of four agents: Emotion Analysis, Emotion Adjustment, Balance Monitoring, and Personal Guide. The Emotion Analysis Agent quantifies stimulus intensity using a 6-emotion BERT classifier, and the Emotion Adjustment Agent rewrites texts into two presentation modes, BALANCED (neutralized text) and COOL (neutralized text + supplementary text), using an LLM. The Balance Monitoring Agent aggregates weekly information consumption patterns and generates personalized advice, while the Personal Guide Agent recommends a presentation mode according to consumer sensitivity. Experiments on 800 AG News articles demonstrated significant stimulus score reduction (up to 19.3\%) and improved emotion balance while maintaining semantic preservation. Near-zero correlation between stimulus reduction and semantic preservation confirmed that the two are independently controllable. Category-level analysis revealed substantial reduction (17.8--33.8\%) in Sports, Business, and Sci/Tech, whereas the effect was limited in the World category, where facts themselves are inherently high-stimulus. The proposed system provides a framework for supporting calm information reception of consumers without restricting access to the original text.
\end{abstract}


\input{Authors/Documents/section1-eg}
\input{Authors/Documents/section3-eg}
\input{Authors/Documents/section4-eg}
\input{Authors/Documents/section5-eg}
\input{Authors/Documents/section6-eg}

\bibliographystyle{IEEEtran}
\bibliography{references}

\vspace{12pt}

\end{document}

%% file: Authors/Documents/section1-eg.tex
\section{Introduction}

In the attention economy, sensational headlines and extreme rhetoric proliferate to capture consumer attention, leading to a near-bulimic consumption of stimulating content~\cite{1}. Such excessive emotional stimulation amplifies anger and anxiety, hindering calm judgment~\cite{2}, and has been linked to stress, social polarization, and the spread of misinformation~\cite{3}. Confirmation bias and filter bubbles in social media further exacerbate this skewed information environment~\cite{4}. Creating an environment in which consumers can engage with information with confidence and make autonomous, calm decisions is therefore an urgent challenge.

Against this backdrop, the concept of informational health has been proposed, drawing an analogy between daily information intake and nutrition and positing that balanced information consumption contributes to the mental well-being of individuals and the soundness of society~\cite{5}. Research on information provision to consumers has traditionally focused on countermeasures against the information itself, such as fake news detection~\cite{6}. In contrast, the present study focuses on the emotional dimension of the information receiver and adjusts how information is conveyed. This approach builds on text style transfer for modifying emotional tone while preserving semantic content~\cite{7,8}, research revealing the association between high-arousal emotions and misinformation spread~\cite{9}, the theory of digital nudging for systematically altering choice behavior through information presentation design~\cite{10}, and multi-agent LLM collaboration through pipeline architectures~\cite{11}.

Building upon these prior studies, this study proposes Multi-Agent LLM-based Emotional deToxification (MALLET), a multi-agent information sanitization system that mitigates and normalizes the emotional impact of information inducing strong emotions such as anger and anxiety, thereby supporting the calm information reception of consumers. The specific contributions are as follows:
\begin{enumerate}
\renewcommand{\labelenumi}{\roman{enumi})}
\item MALLET introduces a pipeline architecture in which four specialized agents operate in coordination, forming a multi-agent LLM system specialized for the emotional protection of consumers.
\item Experimental validation on 800 news texts confirmed both significant stimulus score reduction and semantic preservation, and category-level analysis clarified the applicability and limitations of the proposed method.
\item The Personal Guide Agent and the Balance Monitoring Agent demonstrated that personalization of the presentation mode according to consumer sensitivity and automatic generation of weekly advice based on information consumption patterns are feasible.
\end{enumerate}

The remainder of this paper is organized as follows. Section~II describes the proposed system. Section~III reports the experiments and results. Section~IV discusses the findings and limitations, and Section~V concludes the paper.

%% file: Authors/Documents/section3-eg.tex
\section{Multi-Agent LLM-Based Emotional Detoxification}

\subsection{System Overview}

\begin{figure*}[t]
\centering
\includegraphics[width=0.9\textwidth]{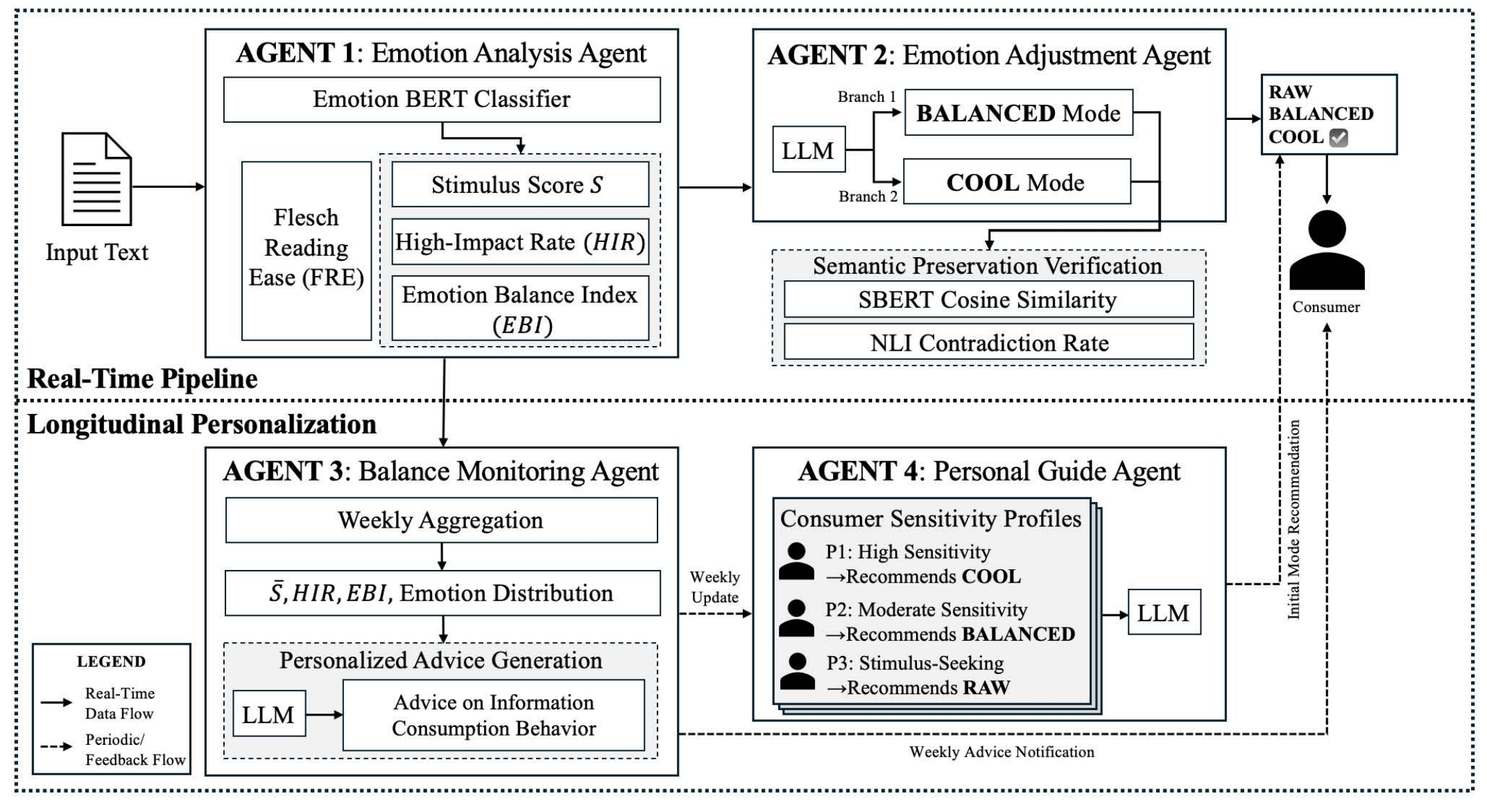}
\caption{Overview of the MALLET System.}
\label{fig:framework_overview}
\end{figure*}

MALLET consists of four specialized AI agents whose outputs are chained sequentially, as shown in Fig.~\ref{fig:framework_overview}. The Emotion Analysis Agent applies a 6-emotion BERT classifier to quantify stimulus intensity as a score in $[0, 1]$. The Emotion Adjustment Agent then rewrites high-stimulus texts into two presentation modes (BALANCED and COOL) using an LLM. The Balance Monitoring Agent aggregates each consumer's browsing logs on a weekly basis and generates personalized advice based on quantitative indicators such as the 6-emotion distribution. The Personal Guide Agent recommends an initial presentation mode (RAW, BALANCED, or COOL) based on the consumer's sensitivity profile. By separating analysis from adjustment, each component can be updated independently, and the two-level presentation modes accommodate diverse consumer needs.

\subsection{Emotion Analysis Agent}

The Emotion Analysis Agent quantifies the emotional tendency of the input text and provides the stimulus intensity criterion to the downstream Emotion Adjustment Agent. Specifically, for each input text $x_i$, it applies a pre-trained 6-emotion BERT model~\cite{12}, to obtain a probability distribution $\mathbf{p}_i = (p_i(c))_{c \in \mathcal{C}}$ over six emotion categories $\mathcal{C} = \{\text{anger},\, \text{fear},\, \text{sadness},\, \text{joy},\, \text{love},\, \text{surprise}\}$. The softmax output of the model is renormalized to sum to 1, and the input is truncated to a maximum of 128 tokens.

From the obtained probability distribution, this study defines three metrics. First, the stimulus score $S_i$, representing the risk level of an individual text, is defined as the sum of the three components---anger, fear, and surprise---which are closely associated with impulsive behavior:
\begin{equation}
S_i = \min\!\bigl(1,\; p_i(\text{anger}) + p_i(\text{fear}) + p_i(\text{surprise})\bigr)
\end{equation}
Second, the High-Impact Rate ($HIR$), representing the risk level at the text-collection level, is defined as the proportion of texts whose stimulus score exceeds a threshold $\theta$:
\begin{equation}
\mathit{HIR} = \frac{1}{N}\sum_{i=1}^{N} \mathbbm{1}[S_i > \theta]
\end{equation}
In this study, $\theta = 0.6$. Third, the Emotion Balance Index ($EBI$), which captures the skewness of the emotion distribution, is defined as the deviation from a uniform distribution:
\begin{equation}
\mathit{EBI}_i = 1 - \frac{1}{2}\sum_{c \in \mathcal{C}} \left| p_i(c) - \frac{1}{6} \right|
\end{equation}
An $\mathit{EBI}$ value close to 1 indicates that the six emotions are evenly distributed, while a low value indicates a bias toward a specific emotion.

In addition, Flesch Reading Ease ($FRE$)~\cite{13}, a widely used readability metric in English, is measured in parallel for all texts to detect increases in text difficulty caused by rewriting.

\subsection{Emotion Adjustment Agent}

The Emotion Adjustment Agent rewrites texts that the Emotion Analysis Agent has identified as high-stimulus into milder expressions while preserving the facts. The rewriting uses GPT-4.1-mini~\cite{14} as the LLM to generate two presentation modes from the original text $x_i$.

The two modes share a common system prompt that instructs the model to ``act as a careful editor, reducing emotional intensity while preserving all facts, proper nouns, numerical values, and core meaning, without adding opinions, and producing concise output.'' On top of this, mode-specific user-level instructions are provided. In BALANCED mode (neutralized text), the model is instructed to replace sensational vocabulary with neutral expressions and output the same amount of information as the original in a single sentence. In COOL mode (neutralized text + supplementary text), the model is instructed to perform the same neutralization as BALANCED and additionally append one supplementary sentence providing background information or definitions to help the reader understand the context calmly, producing a total of two sentences. Both modes are constrained to keep the length within $\pm$20\% of the original. The temperature parameter is set to 0.2 to obtain near-deterministic output that prioritizes fact preservation. The maximum output token count is set to 150.

While rewriting reduces emotional stimulation, it carries the risk of distorting the meaning of the original text. Therefore, two complementary methods are employed for semantic preservation verification. First, both the original text and the rewritten text are encoded using the Sentence-BERT model (SBERT)~\cite{15}, and pairwise cosine similarity is computed. A value closer to 1 indicates higher semantic fidelity. Second, the Natural Language Inference (NLI) model~\cite{16} is given the original text as the premise and the rewritten text as the hypothesis (maximum 256 tokens), and the proportion of pairs for which the argmax among the three classes (entailment, neutral, contradiction) is contradiction is reported as the contradiction rate. A lower contradiction rate indicates that the rewriting has not introduced factual inconsistencies.

\subsection{Balance Monitoring Agent}

While the Emotion Analysis Agent and the Emotion Adjustment Agent reduce stimulus in real time on a per-text basis, the Balance Monitoring Agent assumes the role of overseeing the consumer's information consumption patterns over a longer time horizon. Per-text stimulus reduction alone cannot detect cases in which a consumer chronically continues to consume content biased toward high-stimulus material. To address this challenge, the Balance Monitoring Agent aggregates each consumer's browsing history on a weekly basis and provides data-driven personalized feedback.

Specifically, for the $n_u$ texts that consumer $u$ has viewed during the week, the agent computes a weekly summary consisting of the mean stimulus score $\bar{S}^{(u)}$, $\mathit{HIR}^{(u)}$, $\mathit{EBI}^{(u)}$, and the mean probability for each of the six emotion categories. This enables quantitative visualization of bias toward specific emotions and excessive exposure to high-stimulus content. In the present evaluation, because real consumer logs are not available, 100 texts per consumer are randomly sampled without replacement from a pool of emotion-scored texts to simulate five pseudo-consumers.

The computed weekly summary is converted to JSON format and input to GPT-4.1-mini to generate personalized advice. The model is instructed to act as an expert in diagnosing bias in the consumer's information consumption, to specifically identify which emotions are excessively high or low, and to propose one actionable behavioral change. The generation temperature is set to 0.3. Through this design, abstract biases in numerical indicators are transformed into concrete behavioral suggestions that the consumer can immediately act upon.

\subsection{Personal Guide Agent}

The Emotion Adjustment Agent generates two presentation modes, BALANCED and COOL, but which mode is appropriate for a given consumer depends on individual sensitivity. For consumers who are prone to anxiety, maximum stimulus reduction (COOL) is effective, whereas for consumers who prefer stimulating expressions, retaining the original text (RAW) is preferable. The Personal Guide Agent addresses this individual variation by automatically recommending an initial presentation mode (RAW, BALANCED, or COOL) based on the consumer's psychological profile.

This study defines three consumer personas that span the sensitivity spectrum. Persona~1 is a high-sensitivity type who is sensitive to anxiety and anger and prone to doom-scrolling; Persona~2 is a moderate type who seeks both facts and emotional tone; and Persona~3 is a stimulus-seeking type who enjoys bold expressions while also valuing accuracy. The characteristic description of each persona is input to GPT-4.1-mini, which selects the optimal mode from the three options and outputs a one-sentence justification for the selection. The generation temperature is set to 0.3.

Through this recommendation, consumers can receive information under an appropriate level of protection from their first use without the need for detailed configuration of their sensitivity settings. Furthermore, in practical deployment, the system is designed to iteratively update the recommended mode by referencing the weekly indicators accumulated by the Balance Monitoring Agent, thereby achieving dynamic individual optimization that adapts to changes in the consumer's browsing trends.

%% file: Authors/Documents/section4-eg.tex
\section{Experiments and Results}

\subsection{Experimental Design}

This study uses the AG News dataset~\cite{17}, a standard benchmark for English topic classification. AG News comprises four categories---World, Sports, Business, and Sci/Tech---spanning a wide range of emotional intensities, from high-stimulus conflict reporting to relatively neutral corporate news. A total of 800 texts are obtained by uniformly sampling 200 texts from each class.

The evaluation adopts a within-subject design comparing three conditions, RAW, BALANCED, and COOL, for the same texts. The primary evaluation metrics are the stimulus score $S$, $\mathit{HIR}$ ($\theta=0.6$), $\mathit{EBI}$, $\mathit{FRE}$, and the SBERT cosine similarity and NLI contradiction rate for semantic preservation. Two-tailed paired $t$-tests with Holm correction are used for statistical testing, and effect sizes are evaluated using Cohen's $d$.

\subsection{Evaluation of Stimulus Reduction}

The Emotion Adjustment Agent was applied to the 800 original texts, and the Emotion Analysis Agent was reapplied to each generated mode. The results are shown in Table~\ref{tab:summary}. The mean stimulus score $\bar{S}$ of RAW was 0.482 and $HIR$ was 0.446, indicating that approximately 45\% of original texts exceeded $\theta=0.6$. BALANCED achieved 13.1\% reduction in $\bar{S}$ and 20.6\% in $HIR$, while COOL achieved 19.3\% and 26.0\%, respectively. The greater reduction by COOL is attributed to the supplementary text increasing the proportion of neutral expressions. $EBI$ also improved from 0.316 (RAW) to 0.338 (BALANCED) and 0.346 (COOL), indicating mitigation of emotional bias.

\begin{table}[t]
\centering
\caption{Primary Metrics across Presentation Modes.}
\label{tab:summary}
\begin{tabular}{lcccc}
\toprule
Mode & $\bar{S}$ & HIR & EBI & FRE \\
\midrule
RAW (original) & 0.482 & 0.446 & 0.316 & 39.91 \\
BALANCED (neutralized) & 0.419 & 0.354 & 0.338 & 30.23 \\
COOL (neutralized + suppl.) & 0.389 & 0.330 & 0.346 & 34.54 \\
\bottomrule
\end{tabular}
\end{table}

Readability ($FRE$) decreased to 30.23 for BALANCED (24.3\% decrease) and 34.54 for COOL (13.5\% decrease). BALANCED tends to select more abstract vocabulary under the single-sentence constraint, whereas COOL mitigates this through plain explanations in the supplementary sentence.

Paired $t$-tests with Holm correction confirmed that all six comparisons were significant at $p < 0.001$, as shown in Table~\ref{tab:tests}. The effect sizes for stimulus score reduction were $d=0.210$ (BALANCED) and $d=0.282$ (COOL), both small by Cohen's criteria. The $FRE$ decrease for BALANCED showed a medium effect ($d=0.609$), while COOL was limited to $d=0.339$, confirming the readability compensation effect of the supplementary text.

\begin{table}[t]
\centering
\caption{Results of Paired $t$-Tests (Holm Correction, $n=800$).}
\label{tab:tests}
\begin{tabular}{llccc}
\toprule
Metric & Comparison & $t$ & $p_{\text{adj}}$ & $d$ \\
\midrule
$S$ & RAW vs BAL & 5.93 & $1.3{\times}10^{-8}$ & 0.210 \\
$S$ & RAW vs COOL & 7.97 & $2.1{\times}10^{-14}$ & 0.282 \\
EBI & RAW vs BAL & $-$4.26 & $2.3{\times}10^{-5}$ & $-$0.151 \\
EBI & RAW vs COOL & $-$5.67 & $3.9{\times}10^{-8}$ & $-$0.201 \\
FRE & RAW vs BAL & 17.23 & $5.2{\times}10^{-56}$ & 0.609 \\
FRE & RAW vs COOL & 9.60 & $4.9{\times}10^{-20}$ & 0.339 \\
\bottomrule
\end{tabular}
\end{table}

Category-level analysis revealed clear differences in effectiveness, as shown in Table~\ref{tab:category}. Sports (20.6\%/33.8\%), Business (20.8\%/30.0\%), and Sci/Tech (17.8\%/20.0\%) showed substantial stimulus reduction, whereas the World category showed almost no effect ($-$0.1\% for BALANCED). This is because the World category deals with inherently high-stimulus events such as conflicts and terrorism, where vocabulary such as ``explosion'' and ``attack'' is factually indispensable rather than sensational modification, and thus could not be neutralized at the lexical level. This suggests that alternative intervention strategies such as staged information disclosure are necessary for inherently high-stimulus content.

\begin{table}[t]
\centering
\caption{Mean Stimulus Score by Category.}
\label{tab:category}
\begin{tabular}{lccccc}
\toprule
Category & $\bar{S}_{\text{RAW}}$ & $\bar{S}_{\text{BAL}}$ & $\bar{S}_{\text{COOL}}$ & $\Delta$BAL & $\Delta$COOL \\
\midrule
World & 0.659 & 0.660 & 0.645 & $-$0.1\% & 2.2\% \\
Sports & 0.393 & 0.312 & 0.260 & 20.6\% & 33.8\% \\
Business & 0.500 & 0.396 & 0.350 & 20.8\% & 30.0\% \\
Sci/Tech & 0.374 & 0.308 & 0.299 & 17.8\% & 20.0\% \\
\bottomrule
\end{tabular}
\end{table}


\subsection{Evaluation of Emotion Distribution Changes}

The aggregate metrics used above compress the 6-emotion probability distribution into a single value. This subsection examines individual emotion categories to analyze the mechanism of stimulus reduction.

\begin{figure}[t]
\centering
\includegraphics[width=\columnwidth]{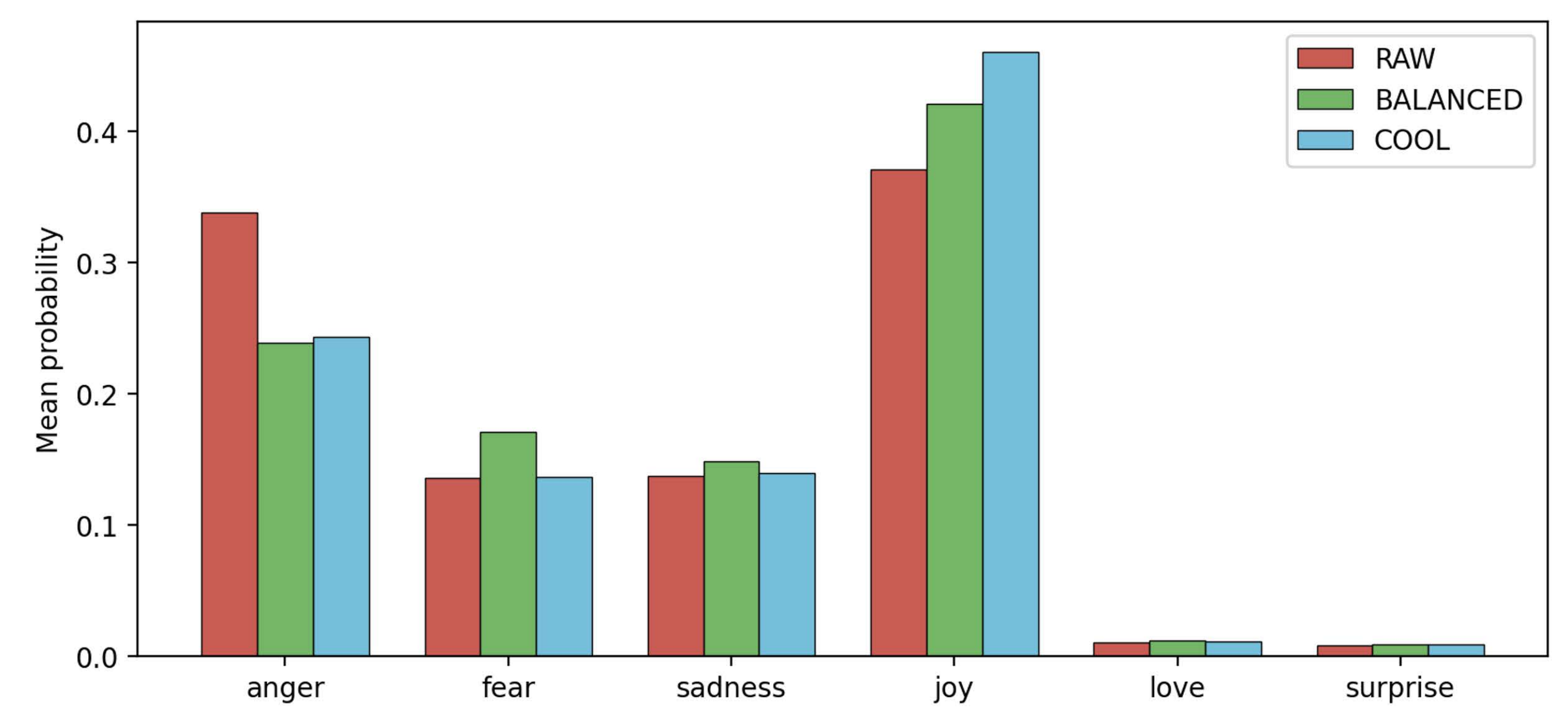}
\caption{Mean Probability Distribution of the Six Emotions across the Three Conditions.}
\label{fig:emotion_dist}
\end{figure}

As shown in Fig.~\ref{fig:emotion_dist}, the most prominent change was the decrease in anger (from 0.34 to 0.24, approximately 29\% reduction) and the increase in joy (from 0.37 to 0.42/0.45). The anger decrease results from the LLM replacing sensational vocabulary (e.g., ``scorching,'' ``outrage'') with neutral expressions (e.g., ``recorded,'' ``reported''). The joy increase is attributable to the softmax constraint: as the anger probability mass decreases, it is redistributed to other categories, with joy receiving the largest share because neutral expressions have features closest to joy in the classifier.

Fear, sadness, love, and surprise showed no substantial changes. The stability of fear is consistent with the limited stimulus reduction in the World category (Section~III-B), confirming that the fear component associated with conflicts and disasters is resistant to lexical neutralization. These results indicate that the primary mechanism of stimulus reduction by MALLET is the selective decrease of the anger component, suggesting that additional interventions targeting fear would be beneficial.

\subsection{Evaluation of Semantic Preservation}

To verify that the rewriting preserved the meaning of the original texts, SBERT cosine similarity and NLI contradiction rate were evaluated, as shown in Table~\ref{tab:semantic}. SBERT similarity was 0.846 for BALANCED and 0.827 for COOL, both indicating high semantic fidelity.

\begin{table}[t]
\centering
\caption{Results of Semantic Preservation Verification.}
\label{tab:semantic}
\begin{tabular}{lcc}
\toprule
Verification method & BALANCED & COOL \\
\midrule
SBERT similarity (mean) & 0.846 & 0.827 \\
NLI entailment rate & 95.1\% & 27.8\% \\
NLI neutral rate & 3.0\% & 68.6\% \\
NLI contradiction rate & 1.9\% & 3.6\% \\
\bottomrule
\end{tabular}
\end{table}

In the NLI judgment distribution, 95.1\% of BALANCED pairs were classified as entailment, while for COOL, neutral accounted for 68.6\%. This is because the supplementary text in COOL was judged as additional information rather than semantic contradiction. The contradiction rate was low for both modes (1.9\% and 3.6\%).

\begin{figure}[t]
\centering
\includegraphics[width=\columnwidth]{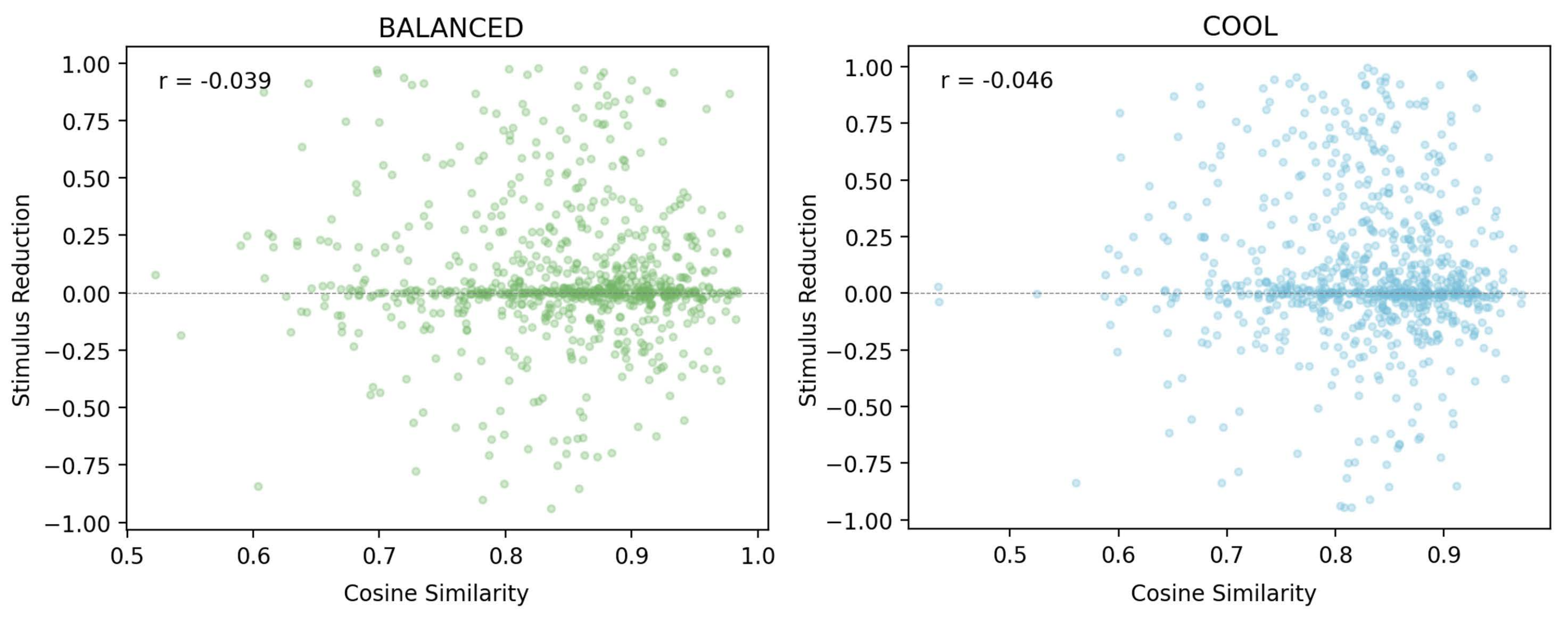}
\caption{Scatter Plot of Stimulus Reduction Magnitude and SBERT Similarity.}
\label{fig:tradeoff}
\end{figure}

The correlation between stimulus reduction magnitude ($S_{\text{RAW}} - S_{\text{mode}}$) and SBERT similarity was $r = -0.039$ (BALANCED) and $r = -0.046$ (COOL), as shown in Fig.~\ref{fig:tradeoff}, indicating near-zero correlation. This confirms that stimulus reduction and semantic preservation are not in a trade-off but are independently achievable.

\subsection{Evaluation of Weekly Aggregation and Personalized Advice}

The Balance Monitoring Agent performed weekly aggregation for five pseudo-consumers, as shown in Table~\ref{tab:users}. $HIR$ ranged from 0.41 to 0.53 and $EBI$ from 0.603 to 0.641, with anger consistently high (0.31--0.40) across all consumers, reflecting the AG News dataset characteristics.

\begin{table}[t]
\centering
\caption{Weekly Summary of Pseudo-Consumers.}
\label{tab:users}
\begin{tabular}{ccccc}
\toprule
Consumer ID & $\bar{S}$ & HIR & EBI & anger \\
\midrule
0 & 0.540 & 0.53 & 0.615 & 0.398 \\
1 & 0.503 & 0.46 & 0.641 & 0.322 \\
2 & 0.537 & 0.50 & 0.609 & 0.364 \\
3 & 0.459 & 0.44 & 0.616 & 0.313 \\
4 & 0.457 & 0.41 & 0.603 & 0.328 \\
\bottomrule
\end{tabular}
\end{table}

Personalized advice was generated based on each consumer's weekly summary. Examples from two contrasting consumers are shown in Table~\ref{tab:advice}. Consumer~0 ($HIR$: 0.53, exceeding 0.5) received advice recommending diversification away from conflict-related content and switching to COOL or BALANCED mode. Consumer~3 (joy-dominant at 0.404, HIR: 0.44, below 0.5) received advice suggesting only topic diversification without mode change. Both pieces of advice contained three elements: i) diagnosis of emotional bias, ii) concrete suggestions for improving information consumption behavior, and iii) numerical justification. This confirmed that abstract weekly indicators can be translated into actionable improvements in information consumption patterns.

\begin{table}[t]
\centering
\caption{Examples of Personalized Advice.}
\label{tab:advice}
\begin{tabular}{cp{6cm}}
\toprule
ID & Advice \\
\midrule
0 & Your dominant emotion this week is anger at 39.8\%, combined with a high HIR of 0.53, indicating frequent engagement with conflict-driven news. Try diversifying your news consumption toward science, technology, or positive stories. Additionally, switching your display mode to COOL or BALANCED can help moderate emotional impact. \\
3 & Your dominant emotion this week is joy (0.404), with moderate anger at 0.313 and HIR of 0.44. To maintain a positive balance while reducing anger, try incorporating more science and technology stories. Since your HIR is below 0.5, you can continue with your current display mode but consider mixing in more balanced topics. \\
\bottomrule
\end{tabular}
\end{table}


\subsection{Evaluation of Mode Recommendation}

The Personal Guide Agent performed mode recommendation for three consumer personas, as shown in Table~\ref{tab:guide}. A graded recommendation of COOL $\to$ BALANCED $\to$ RAW was made in order of decreasing sensitivity. This graded nature is consistent with the stimulus reduction demonstrated in Section~III-B (COOL: 19.3\% $>$ BALANCED: 13.1\% $>$ RAW: 0\%), confirming that adaptive personalization of information presentation according to consumer sensitivity is achievable.

\begin{table}[t]
\centering
\caption{Mode Recommendation Results Based on Consumer Personas.}
\label{tab:guide}
\begin{tabular}{clp{4.5cm}}
\toprule
ID & Rec. & Rationale \\
\midrule
P1 & COOL & Gentle and context-rich rewriting reduces anxiety and prevents being overwhelmed. \\
P2 & BAL & Compatibility between factual accuracy and moderate emotional tone suits a balanced consumer. \\
P3 & RAW & Preserves the boldness and dramatic expression of the original while maintaining accuracy, matching the stimulus-seeking preference. \\
\bottomrule
\end{tabular}
\end{table}

%% file: Authors/Documents/section5-eg.tex
\section{Discussion}

The novelty of this study lies in demonstrating that the intensity of emotional stimulation can be controlled in a graded manner, in contrast to prior text style transfer studies~\cite{7, 8}, which aimed at polarity conversion of emotions. The two-level adjustment through BALANCED and COOL achieved stimulus reduction of 13.1\% and 19.3\%, respectively, while the near-zero correlation observed between the degree of stimulus reduction and semantic preservation indicates that the two are not in a trade-off but are independently controllable. Furthermore, category-level analysis revealed that this effect is attributable to the removal of sensational modification, and simultaneously demonstrated the applicability limit that lexical-level neutralization does not function for the World category, where the facts themselves are inherently high-stimulus. Considering the association between high-arousal emotions and misinformation demonstrated by the fake news detection of Zhang et al.~\cite{9}, this finding suggests the need for a consumer protection framework that complementarily combines detection (post-hoc) and mitigation (pre-emptive).

On the other hand, this study has several limitations. The evaluation of stimulus reduction depends on the output of the emotion classifier, and verification through human subjective emotional judgment is necessary. The readability decrease in BALANCED mode ($d=0.609$) requires the explicit incorporation of readability constraints in the prompt design. The evaluation of the Balance Monitoring Agent is based on pseudo-consumers, and longitudinal verification using real consumer browsing logs remains a task for future work. Additionally, the study targets only English texts, and the applicability to multilingual contexts including Japanese has not been verified.

%% file: Authors/Documents/section6-eg.tex
\section{Conclusion}

This study proposed MALLET, a multi-agent information sanitization system consisting of four AI agents. Through experiments using 800 news texts, both significant stimulus score reduction and semantic preservation were demonstrated, and it was shown that the two are not in a trade-off but are independently controllable. Category-level analysis revealed the effectiveness of the system for news containing sensational modification and its limitations for content where the facts themselves are inherently high-stimulus. The Balance Monitoring Agent and the Personal Guide Agent confirmed that personalization according to the consumer's information consumption patterns and sensitivity is feasible. 

Future work includes verification through human evaluation, alternative interventions for inherently high-stimulus content, longitudinal evaluation using real consumer data, and multilingual support.